\documentclass[11pt,a4paper]{article}
\usepackage[hyperref]{acl2020}
\usepackage{times}
\usepackage{latexsym}

\usepackage{microtype}

\aclfinalcopy 

\usepackage{graphicx}
\usepackage{amsmath}
\usepackage{dsfont}
\usepackage{amsfonts}
\usepackage{mathtools}
\usepackage{multirow}
\usepackage{subcaption,siunitx,booktabs}
\usepackage{enumitem}
\usepackage{CJKutf8}
\usepackage{xspace}

\usepackage{cleveref}
\crefname{section}{\S}{\S\S}
\crefname{table}{Table}{}
\crefname{figure}{Figure}{}
\crefname{algorithm}{Algorithm}{}
\crefname{equation}{eq.}{}
\crefname{appendix}{App.}{}
\crefname{prop}{Proposition}{}
\crefname{thm}{Theorem}{}
\crefformat{section}{\S#2#1#3}  

\newcommand{\bleu}{\textsc{Bleu}\xspace}
\newcommand{\ribes}{\textsc{Ribes}\xspace}
\newcommand{\pascal}{\textsc{Pascal}\xspace}
\newcommand{\lisa}{\textsc{LISA}\xspace}
\newcommand{\multi}{\textsc{Multi-Task}\xspace}
\newcommand{\sh}{\textsc{S$\text{\&}$H}\xspace}

\usepackage{todonotes}
\makeatletter
\newcommand*\iftodonotes{\if@todonotes@disabled\expandafter\@secondoftwo\else\expandafter\@firstoftwo\fi}  
\makeatother

\title{Enhancing Machine Translation with Dependency-Aware Self-Attention}

\author{Emanuele Bugliarello\thanks{~~Work done while at Tokyo Institute of Technology.}\\University of Copenhagen\\\texttt{emanuele@di.ku.dk}
        \And 
        Naoaki Okazaki\\Tokyo Institute of Technology\\\texttt{okazaki@c.titech.ac.jp}
       }

\date{}

\begin{document}
\maketitle
\begin{abstract}
Most neural machine translation models only rely on pairs of parallel sentences, assuming syntactic information is automatically learned by an attention mechanism.
In this work, we investigate different approaches to incorporate syntactic knowledge in the Transformer model and also propose a novel, parameter-free, dependency-aware self-attention mechanism that improves its translation quality, especially for long sentences and in low-resource scenarios.
We show the efficacy of each approach on WMT English$\leftrightarrow$German and English$\rightarrow$Turkish, and WAT English$\rightarrow$Japanese translation tasks.
\end{abstract}

\section{Introduction}

\begin{figure*}[t]
	\centering
	\includegraphics[width=\textwidth, trim={0.5cm 3.5cm .5cm 0.cm}, clip]{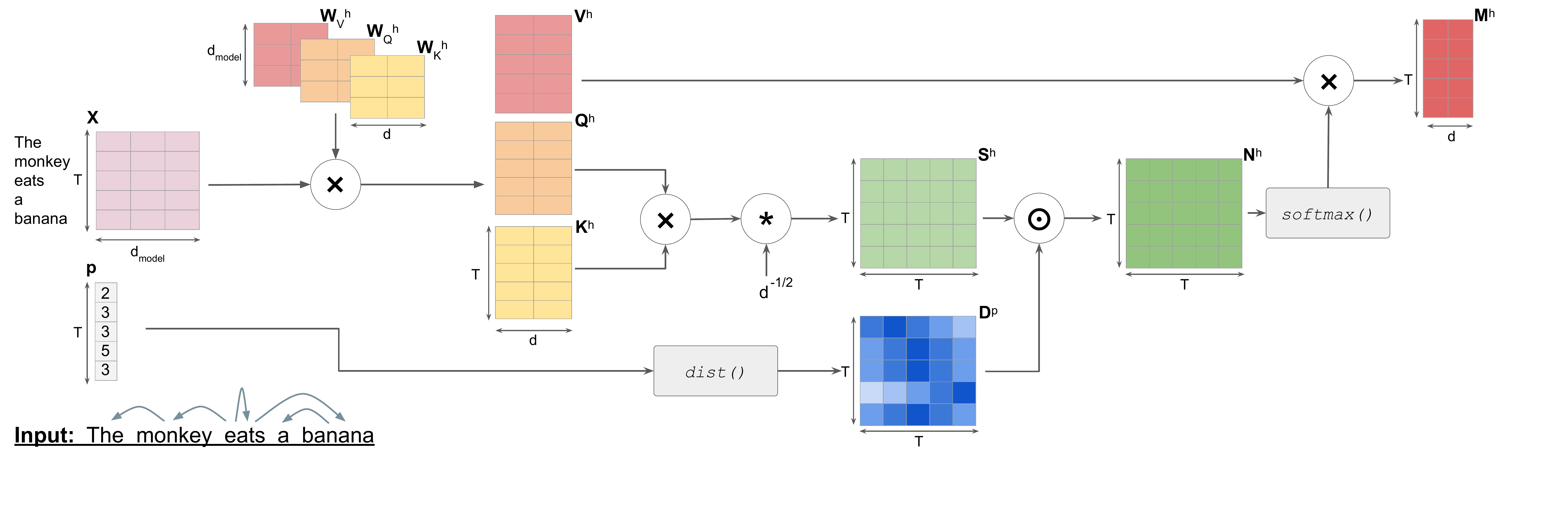}
	\caption{Parent-Scaled Self-Attention (\pascal) head for the input sequence ``The monkey eats a banana".}
	\label{fig:pascal}
\end{figure*}

Research in neural machine translation (NMT) has mostly exploited corpora consisting of pairs of parallel sentences, with the assumption that a model can automatically learn prior linguistic knowledge via an attention mechanism~\cite{luong2015effective}.
However, \newcite{shi-etal-2006-dom} found that these models still fail to capture deep structural details, and several studies~\cite{sennrich2016linguistic,P17-2012,P17-1177,AAAI1816060} have shown that syntactic information has the potential to improve these models.
Nevertheless, the majority of syntax-aware NMT models are based on recurrent neural networks (RNNs;~\citealt{elman1990finding}), with only a few recent studies that have investigated methods for the Transformer model~\cite{vaswani2017attention}.

\newcite{wu2018dep2dep} evaluated an approach to incorporate syntax in NMT with a Transformer model, which not only required three encoders and two decoders, but also target-side dependency relations (precluding its use to low-resource target languages).
\newcite{zhang-etal-2019-syntax} integrate source-side syntax by concatenating the intermediate representations of a dependency parser to word embeddings.
In contrast to ours, this approach does not allow to learn sub-word units at the source side, requiring a larger vocabulary to minimize out-of-vocabulary words.
\newcite{saunders-etal-2018-multi} interleave words with syntax representations which results in longer sequences -- requiring gradient accumulation for effective training -- while only leading to $+0.5$ \bleu on WAT Ja-En when using ensembles of Transformers.
Finally, \newcite{currey-heafield-2019-incorporating} propose two simple data augmentation techniques to incorporate source-side syntax: one that works well on low-resource data, and one that achieves a high score on a large-scale task.
Our approach, on the other hand, performs equally well in both settings.

While these studies improve the translation quality of the Transformer, they do not exploit its properties. 
In response, we propose to explicitly enhance the its self-attention mechanism (a core component of this architecture) to include syntactic information without compromising its flexibility.
Recent studies have, in fact, shown that self-attention networks benefit from modeling local contexts by reducing the dispersion of the attention distribution~\cite{shaw-etal-2018-self,yang-etal-2018-modeling,Yang2019ContextAwareSN}, and that they might not capture the inherent syntactic structure of languages as well as recurrent models, especially in low-resource settings~\cite{tran-etal-2018-importance,tang-etal-2018-self}. 
Here, we present \textit{parent-scaled self-attention} (\pascal): a novel, parameter-free local attention mechanism that lets the model focus on the dependency parent of each token when encoding the source sentence.
Our method is simple yet effective, improving translation quality with no additional parameter or computational overhead.

Our main contributions are:
\begin{itemize}[noitemsep,topsep=1pt]
    \item introducing \pascal: an effective parameter-free local self-attention mechanism to incorporate source-side syntax into Transformers;
    \item adapting \lisa~\cite{D18-1548} to sub-word representations and applying it to NMT;
    \item similar to concurrent work~\cite{pham2019promoting}, we find that modeling linguistic knowledge into the self-attention mechanism leads to better translations than other approaches. 
\end{itemize}
Our extensive experiments on standard En$\leftrightarrow$De, En$\rightarrow$Tr and En$\rightarrow$Ja translation tasks also show that (a) approaches to embed syntax in RNNs do not always transfer to the Transformer, and (b) \pascal consistently exhibits significant improvements in translation quality, especially for long sentences.

\section{Model}\label{sec:model}
In order to design a neural network that is efficient to train and that exploits syntactic information while producing high-quality translations, we base our model on the Transformer architecture \cite{vaswani2017attention} and upgrade its encoder with \textit{parent-scaled self-attention} (\pascal) heads at layer $l_s$.
\pascal heads enforce contextualization from the syntactic dependencies of each source token, and, in practice, we replace standard self-attention heads with \pascal ones in the first layer as its inputs are word embeddings that lack any contextual information. 
Our \pascal sub-layer has the same number $H$ of attention heads as other layers.

\paragraph{Source syntax}
Similar  to  previous  work,  instead  of  just  providing  sequences of tokens, we supply the encoder with dependency relations given by an external parser. 
Our approach explicitly exploits sub-word units, which enable open-vocabulary translation: after generating sub-word units, we compute the middle position of each word in terms of number of tokens. For instance, if a word in position $4$ is split into three tokens, now in positions $6$, $7$ and $8$, its middle position is $7$. 
We then map each sub-word of a given word to the middle position of its parent. For the root word, we define its parent to be itself, resulting in a parse that is a directed graph. 
The input to our encoder is a sequence of $T$ tokens and the absolute positions of their parents.

\subsection{Parent-Scaled Self-Attention} \label{sec:pascal}

Figure~\ref{fig:pascal} shows our parent-scaled self-attention sub-layer. 
Here, for a sequence of length $T$, the input to each head is a matrix $\mathbf{X}\in\mathbb{R}^{T\times d_{model}}$ of token embeddings and a vector $\textbf{p}\in\mathbb{R}^T$ whose $t$-th entry $p_t$ is the middle position of the $t$-th token's dependency parent.
Following~\newcite{vaswani2017attention}, in each attention head $h$, we compute three vectors (called query, key and value) for each token, resulting in the three matrices $\mathbf{K}^h\in\mathbb{R}^{T\times d}$, $\mathbf{Q}^h\in\mathbb{R}^{T\times d}$, and $\mathbf{V}^h\in\mathbb{R}^{T\times d}$ for the whole sequence, where $d = d_{model}/H$.
We then compute dot products between each query and all the keys, giving scores of how much focus to place on other parts of the input when encoding a token at a given position.
The scores are divided by $\sqrt{d}$ to alleviate the vanishing gradient problem arising if dot products are large:
\begin{equation}
\mathbf{S}^h = \mathbf{Q}^h~{\mathbf{K}^h}^\top / \sqrt{d}.
\end{equation}
Our main contribution is in weighing the scores of the token at position $t$, $\textbf{s}_t$, by the distance of each token from the position of $t$'s dependency parent:
\begin{equation}\label{eq:scaling}
n^h_{tj} = s^h_{tj} ~ d^p_{tj}, ~~\text{ for } j=1,...,T,
\end{equation}
where $\textbf{n}^h_{t}$ is the $t$-th row of the matrix $\mathbf{N}^h\in\mathbb{R}^{T\times T}$ representing scores normalized by the proximity to $t$'s parent. $d^p_{tj} = dist(p_t, j)$ is the $\left(t, j\right)^{th}$ entry of the matrix $\mathbf{D}^p\in\mathbb{R}^{T\times T}$ containing, for each row $\mathbf{d}_t$, the distances of every token $j$ from the middle position of token $t$'s dependency parent $p_t$.
In this paper, we 
compute this distance as the value of the probability density of a normal distribution centered at $p_t$ and with variance $\sigma^2$, $\mathcal{N}\left(p_t,\sigma^{2}\right)$:
\begin{equation}
dist(p_t, j) = f_{\mathcal{N}}\left(j\middle\vert p_t, \sigma^2\right) = \frac{1}{\sqrt{2\pi\sigma^2}}e^{-\frac{\left(j - p_t\right)^2}{2\sigma^2}}.
\end{equation}

Finally, we apply a softmax function to yield a distribution of weights for each token over all the tokens in the sentence, and multiply the resulting matrix with the value matrix $\textbf{V}^h$, obtaining the final representations $\mathbf{M}^h$ for \pascal head $h$.

One of the major strengths of our proposal is being parameter-free: no additional parameter is required to train our \pascal sub-layer as $\textbf{D}^p$ is obtained by computing a distance function that only depends on the vector of tokens' parent positions and can be evaluated using fast matrix operations.


\paragraph{Parent ignoring}
Due to the lack of parallel corpora with gold-standard parses, we rely on noisy annotations from an external parser. However, the performance of syntactic parsers drops abruptly when evaluated on out-of-domain data~\cite{dredze2007frustratingly}.
To prevent our model from overfitting to noisy dependencies, we introduce a regularization technique for the \pascal sub-layer: \textit{parent ignoring}. 
In a similar vein as dropout~\cite{JMLR:v15:srivastava14a}, we disregard information during the training phase. Here, we ignore the position of the parent of a given token by randomly setting each row of $\mathbf{D}^p$ to $\mathbf{1}\in\mathbb{R}^T$ with some probability $q$.


\paragraph{Gaussian weighing function}
The choice of weighing each score by a Gaussian probability density is motivated by two of its properties. 
First, its bell-shaped curve: It allows us to focus most of the probability density at the mean of the distribution, which we set to the middle position of the sub-word units of the dependency parent of each token.
In our experiments, we find that most words in the vocabularies are not split into sub-words, hence allowing \pascal to mostly focus on the actual parent. In addition, non-negligible weights are placed on the neighbors of the parent token, allowing the  attention mechanism to also attend to them. This could be useful, for instance, to learn idiomatic expressions such as prepositional verbs in English.
The second property of Gaussian-like distributions that we exploit is their support: While most of the weight is placed in a small window of tokens around the mean of the distribution, all the values in the sequence are actually multiplied by non-zero factors; allowing a token $j$ farther away from the parent of token $t$, $p_t$, to still play a role in the representation of $t$ if its score $s_{tj}^h$ is high.

\paragraph{}
\pascal can be seen as an extension of the local attention mechanism of~\newcite{luong2015effective}, with the alignment now guided by syntactic information. 
\newcite{yang-etal-2018-modeling} proposed a method to learn a Gaussian bias that is added to, instead of multiplied by, the original attention distribution.
As we will see next, our model significantly outperforms this.

\section{Experiments}
\begin{figure*}[t]
	\center
	\includegraphics[width=\textwidth, trim={0cm 0.2cm 0cm 0.3cm}, clip]{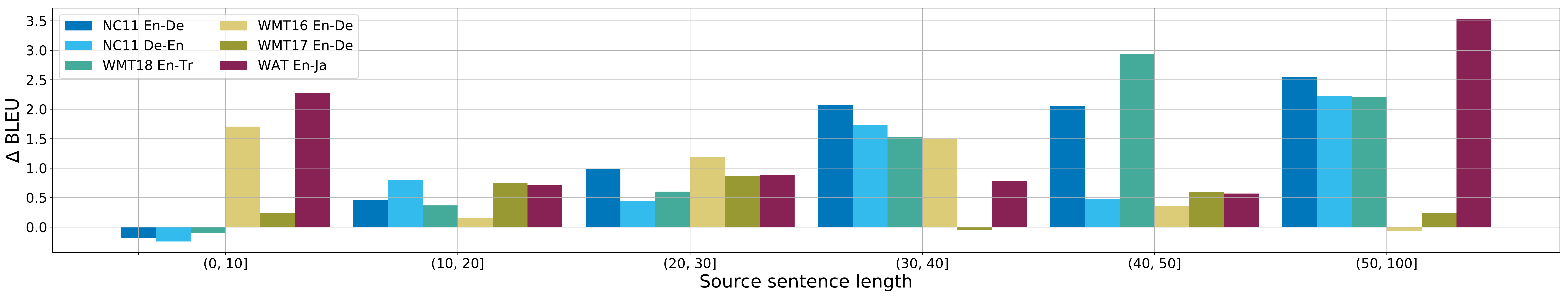}
	\includegraphics[width=\textwidth, trim={0cm 0.2cm 0cm 0.3cm}, clip]{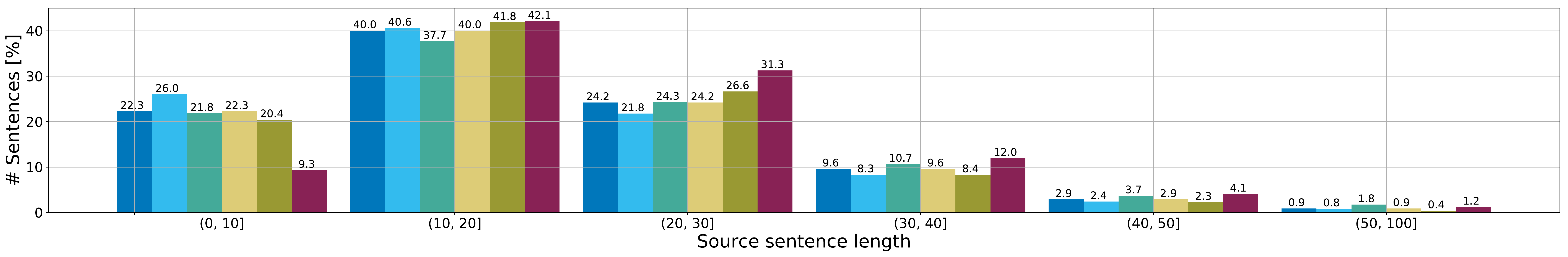}
	\vspace{-20pt}
	\caption{Analysis by sentence length: $\Delta$\bleu with the Transformer (above) and percentage of data (below).}
	\label{fig:len2bleu}
\end{figure*}
\begin{table*}[t]
    \center
    \small
    \setlength\tabcolsep{5pt}
    \begin{minipage}{\textwidth} 
    \center
    \begin{tabular}{l|r|r|r||r|r|r|r}
        \toprule
        \multirow{2}{*}{\textbf{Method}} 
            & \textbf{NC11}  & \textbf{NC11}  & \textbf{WMT18} & \textbf{WMT16} & \textbf{WMT17} & \multicolumn{2}{c}{\textbf{WAT}} \\
            & \textbf{En-De} & \textbf{De-En} & \textbf{En-Tr} & \textbf{En-De} & \textbf{En-De} & \textbf{En-Ja} [B] & \textbf{En-Ja} [R] \\
        \midrule
        \newcite{eriguchi2016tree} & & & & & & 34.9 & 81.58 \\
        \newcite{D17-1209} & 16.1 & & & & &\\
        \newcite{D17-1012} & & & & & & 39.4 & 82.83 \\
        \newcite{tran2018inducing} & & & & 30.3 & 24.3 & &\\
    	SE+SD-NMT$^\dagger$~\cite{wu2018dep2dep} & & & & & 24.7 & 36.4 & 81.83 \\
        \ 
        SE+SD-Transformer$^\dagger$~\cite{wu2018dep2dep} & & & & & \textbf{26.2} & &\\
        Mixed Enc.~\cite{currey-heafield-2019-incorporating} & & & 9.6 & 31.9 & 26.0 & &\\
        Multi-Task~\cite{currey-heafield-2019-incorporating} & & & 10.6 & 29.6 & 23.4 & &\\
        \hline
        Transformer & 25.0 & 26.6 & 13.1 & 33.0 & 25.5 & 43.1 & 83.46\\
        ~~~~+ \pascal & \textbf{25.9}$^\Uparrow$ & \textbf{27.4}$^\Uparrow$ & \textbf{14.0}$^\Uparrow$ & \textbf{33.9}$^\Uparrow$ & \textit{26.1}$^\Uparrow$ & \textbf{44.0}$^\Uparrow$ & \textbf{85.21}$^\Uparrow$ \\ 
        ~~~~+ \lisa & 25.3 & 27.1 & 13.6 & 33.6 & 25.7 & 43.2 & 83.51 \\
        ~~~~+ \multi & 24.8 & 26.7 & \textbf{14.0} & 32.4 & 24.6 & 42.7 & 84.18 \\
    	~~~~+ \sh & 25.5 & 26.8 & 13.0 & 31.9 & 25.1 & 42.8 & 83.88 \\
    	\bottomrule
    \end{tabular}
    \end{minipage}
    \vspace{-5pt}
    \caption{Test \bleu (and \ribes for En-Ja) scores on small-scale (left) and large-scale (right) data sets. Models that also require target-side syntax information are marked with $^\dagger$, while $^\Uparrow$ indicates statistical significance ($p < 0.01$) against the Transformer baseline via bootstrap re-sampling~\cite{koehn-2004-statistical}.}\label{tab:results}
\end{table*}
\subsection{Experimental Setup}
\paragraph{Data} 

We evaluate the efficacy of our approach on standard, large-scale benchmarks and on low-resource scenarios, where the Transformer was shown to induce poorer syntax. 
Following~\newcite{D17-1209}, we use News Commentary v11 (NC11) with En-De and De-En tasks to simulate low resources and test multiple source languages.
To compare with previous work, we train our models on WMT16 En-De and WAT En-Ja tasks, removing sentences in incorrect languages from WMT16 data sets.
For a thorough comparison with concurrent work, we also evaluate on the large-scale WMT17 En-De and low-resource WMT18 En-Tr tasks.
We rely on Stanford CoreNLP~\cite{manning-EtAl:2014:P14-5} to parse source sentences.\footnote{\label{ft:app}For a detailed description, see Appendix~\ref{sec:full_exp}.}

\paragraph{Training}
We implement our models in PyTorch on top of the Fairseq toolkit.\footnote{\url{https://github.com/e-bug/pascal}.}
Hyperparameters, including the number of \pascal heads, that achieved the highest validation \bleu~\cite{Papineni:2002:BMA:1073083.1073135} score were selected via a small grid search. 

We report previous results in syntax-aware NMT for completeness, and train a Transformer model as a strong, standard baseline. We also investigate the following syntax-aware Transformer approaches:\textsuperscript{\ref{ft:app}}
\begin{itemize}[noitemsep,topsep=1pt]
    \item \textbf{+\pascal:} The model presented in \cref{sec:model}.
    The variance of the normal distribution was set to $1$ (i.e., an effective window size of $3$) as $99.99\%$ of the source words in our training sets are at most split into $7$ sub-words units.\\[-12pt]
    \item \textbf{+\lisa:} We adapt LISA~\cite{D18-1548} to NMT and sub-word units by defining the parent of a given token as its first sub-word (which represents the root of the parent word).\\[-12pt]
    \item \textbf{+\multi:} Our implementation of the multi-task approach by~\newcite{currey-heafield-2019-incorporating} where a standard Transformer learns to both parse and translate source sentences.\\[-12pt]
    \item \textbf{+\sh:} Following~\newcite{sennrich2016linguistic}, we introduce syntactic information in the form of dependency labels in the embedding matrix of the Transformer encoder.
\end{itemize}

\subsection{Results}
Table~\ref{tab:results} presents the main results of our experiments. 
Clearly, the base Transformer outperforms previous syntax-aware RNN-based approaches, proving it to be a strong baseline in our experiments.
The table shows that the simple approach of~\newcite{sennrich2016linguistic} does not lead to notable advantages when applied to the embeddings of the Transformer model.
We also see that the multi-task approach benefits from better parameterization, but it only attains comparable performance with the baseline on most tasks.
On the other hand, \lisa, which embeds syntax in a self-attention head, leads to modest but consistent gains across all tasks, proving that it is also useful for NMT.
Finally, \pascal outperforms all other methods, with consistent gains over the Transformer baseline independently of the source language and corpus size: It gains up to $+0.9$ \bleu points on most tasks and a substantial $+1.75$ in \ribes~\cite{isozaki2010ribes}, a metric with stronger correlation with human judgments than \bleu in En$\leftrightarrow$Ja translations.
On WMT17, our slim model compares favorably to other methods, achieving the highest \bleu score across all source-side syntax-aware approaches.\footnote{Note that modest improvements in this task should not be surprising as Transformers learn better syntactic relationships from larger data sets~\cite{raganato-tiedemann-2018-analysis}.}

Overall, our model achieves substantial gains given the grammatically rigorous structure of English and German. 
Not only do we expect performance gains to further increase on less rigorous sources and with better parses~\cite{zhang-etal-2019-syntax}, but also higher robustness to noisier syntax trees obtained from back-translated with parent ignoring.

\paragraph{Performance by sentence length}
As shown in Figure~\ref{fig:len2bleu}, our model is particularly useful when translating long sentences, obtaining more than $+2$ \bleu points when translating long sentences in all low-resource experiments, and $+3.5$ \bleu points on the distant En-Ja pair.
However, only a few sentences ($1\%$) in the evaluation datasets are long.

\begin{table}[t]
	\centering
	\small
	\setlength\tabcolsep{2pt}
    \begin{CJK}{UTF8}{min}
	\begin{tabular}{l|l}
		\toprule
        \textbf{SRC} & In a cooling experiment , \textbf{only} a tendency agreed \\
        \textbf{BASE} & 冷却 実験 で は ，\textbf{わずか な} 傾向 が 一致 し た \\
        \textbf{OURS} & 冷却 実験 で は 傾向 \textbf{のみ} 一致 し た \\
        \midrule
        \textbf{SRC} & Of course I \textbf{don't} hate you \\
        \textbf{BASE} & Natürlich hass\textbf{te} ich dich nicht \\
        \textbf{OURS} & Natürlich hass\textbf{e} ich dich nicht \\
        \midrule
        \textbf{SRC} & What are those people fighting for? \\
        \textbf{BASE} & Was sind die Menschen, für die kämpfen? \\
        \textbf{OURS} & Wofür kämpfen diese Menschen? \\
		\bottomrule
	\end{tabular}
	\end{CJK}
	\caption{Example of correct translation by Pascal.} \label{tab:example}
\end{table}
\paragraph{Qualitative performance}
Table~\ref{tab:example} presents examples where our model correctly translated the source sentence while the Transformer baseline made a syntactic error.
For instance, in the first example, the Transformer misinterprets the adverb ``only" as an adjective of ``tendency:'' the word ``only'' is an adverb modifying the verb ``agreed.''
In the second example, ``don't'' is incorrectly translated to the past tense instead of present.

\paragraph{\pascal layer}
When we introduced our model, we motivated our design choice of placing \pascal heads in the first layer in order to enrich the representations of words from their isolated embeddings by introducing contextualization from their parents.
We ran an ablation study on the NC11 data in order to verify our hypothesis.
As shown in Table~\ref{tab:pascal_layer}, the performance of our model on the validation sets is lower when placing Pascal heads in upper layers; a trend that we also observed with the \lisa mechanism. 
These results corroborate the findings of~\newcite{raganato-tiedemann-2018-analysis} who noticed that, in the first layer, more attention heads solely focus on the word to be translated itself rather than its context.
We can then deduce that enforcing syntactic dependencies in the first layer effectively leads to better word representations, which further enhance the translation accuracy of the Transformer model.
Investigating the performance of multiple syntax-aware layers is left as future work.

\paragraph{Gaussian variance}
Another design choice we made was the variance of the Gaussian weighing function.
We set it to $1$ in our experiments motivated by the statistics of our datasets, where the vast majority of words is at most split into a few tokens after applying BPE.
Table~\ref{tab:var} corroborates our choice, showing higher \bleu scores on the NC11 validation sets when the variance equals $1$.
Here, ``parent-only" is the case where weights are only placed to the middle token (i.e. the parent).

\begin{table}[t]
    \begin{subtable}{.5\linewidth}
        \centering
        \small
        \setlength\tabcolsep{3pt}
        \begin{tabular}{l|r|r}
    		\toprule
    		\textbf{Layer} & \textbf{En-De} & \textbf{De-En} \\ 
    		\midrule 
    		1 & \textbf{23.2} & \textbf{24.6} \\
    		2 & 22.5 & 20.1 \\
            3 & 22.5 & 23.8 \\
            4 & 22.6 & 23.8 \\
            5 & 22.9 & 23.8 \\
            6 & 22.4 & 23.9 \\
    		\bottomrule
    	\end{tabular}
    	\caption{}\label{tab:pascal_layer}
    \end{subtable}%
    \begin{subtable}{.5\linewidth}
        \centering
        \small
        \setlength\tabcolsep{3pt}
        \begin{tabular}{l|r|r}
    		\toprule
    		\textbf{Variance} & \textbf{En-De} & \textbf{De-En} \\ 
    		\midrule 
    		Parent-only & 22.5 & 22.4 \\
    		1 & \textbf{23.2} & \textbf{24.6} \\
            4 & 22.7 & 24.3 \\
            9 & 22.8 & 24.3 \\
            16 & 22.7 & 24.4 \\
            25 & 22.8 & 24.1 \\
    		\bottomrule
    	\end{tabular}
    \caption{}\label{tab:var}
    \end{subtable} 
    \caption{Validation \bleu as a function of \pascal layer (a) and Gaussian's variance (b) on NC11 data.}
\end{table}

\paragraph{Sensitivity to hyperparameters}
Due to the large computational cost required to train Transformer models, we only searched hyperparameters in a small grid.
In order to estimate the sensitivity of the proposed approach to hyperparameters, we trained the NC11 De-En model with the hyperparameters of the En-De one. 
In fact, despite being trained on the same data set, we find that more \pascal heads help when German (which has a higher syntactic complexity than English) is used as the source language.
In this test, we only find $-0.2$ \bleu points with respect to the score listed in Table~\ref{tab:results}, showing that our general approach is effective regardless of extensive fine-tuning.

Additional analyses are reported in Appendix~\ref{sec:analysis}.

\section{Conclusion}
This study provides a thorough investigation of approaches to induce syntactic knowledge into self-attention networks.
Through extensive evaluations on various translation tasks, we find that approaches effective for RNNs do not necessarily transfer to Transformers (e.g. +\sh).
Conversely, dependency-aware self-attention mechanisms (\lisa and \pascal) best embed syntax, for all corpus sizes, with \pascal consistently outperforming other all approaches.
Our results show that exploiting core components of the Transformer to embed linguistic knowledge leads to higher and consistent gains than previous approaches. 

\section*{Acknowledgments}
We are grateful to the anonymous reviewers, Desmond Elliott and the CoAStaL NLP group for their constructive feedback.
The research results have been achieved by ``Research and Development of Deep Learning Technology for Advanced Multilingual Speech Translation,'' the Commissioned Research of National Institute of Information and Communications Technology (NICT), Japan.

\bibliography{acl2020}
\bibliographystyle{acl_natbib}

\clearpage
\appendix

\section{Experiment details}\label{sec:full_exp}
\begin{table}
	\centering
	\small
		\setlength\tabcolsep{3pt}
	\begin{tabular}{l|r|r|r|r}
		\toprule
		\multicolumn{1}{c|}{\textbf{Corpus}} & \multicolumn{1}{c|}{\textbf{Train}} & \multicolumn{1}{c|}{\textbf{Filtered Train}} & \multicolumn{1}{c|}{\textbf{Valid}} & \multicolumn{1}{c}{\textbf{Test}} \\
		\midrule
		NC11~~~~ En-De & 238,843 & 233,483 & 2,169 & 2,999 \\
		WMT18 En-Tr & 207,373 &  & 3,000 & 3,007 \\\hline
		WMT16 En-De & 4,500,962 & 4,281,379 & 2,169 & 2,999 \\
		WMT17 En-De & 5,852,458 &  & 2,999 & 3,004 \\
		WAT~~~~~~~En-Ja & 3,008,500 &  & 1,790 & 1,812 \\
		\bottomrule
	\end{tabular}
	\caption{Number of sentences in each data set.} \label{tab:datasets}
\end{table}

\paragraph{Data preparation}
We follow the same pre-processing steps as~\newcite{vaswani2017attention}.
Unless otherwise specified, we first tokenize the data with Moses~\cite{Koehn:2007:MOS:1557769.1557821} and remove sentences longer than $80$ tokens in either source or target side.

Following~\newcite{D17-1209}, we train on News Commentary v11 (NC11) data set 
with English$\rightarrow$German (En-De) and German$\rightarrow$English (De-En) tasks so as to simulate low-resource cases and to evaluate the performance of our models for different source languages.
We also train on the full WMT16 data set for En-De, using \textit{newstest2015} and \textit{newstest2016} as validation and test sets, respectively, in each of these experiments.
Moreover, we notice that these data sets contain sentences in different languages and use \texttt{langdetect}\footnote{\url{https://pypi.org/project/langdetect}.} to remove sentences
in incorrect languages. 

We also train our models on WMT18 
English$\rightarrow$Turkish (En-Tr) as a standard low-resource scenario.
Models are evaluated on \textit{newstest2016} and tested on \textit{newstest2017}.

Previous studies on syntax-aware NMT have commonly been conducted on the WMT16 En-De and WAT English$\rightarrow$Japanese (En-Ja) tasks, while concurrent approaches are evaluated on the WMT17 
En-De task.
In order to provide a generic and comprehensive evaluation of our proposed approach on large-scale data, we also train our models on the latter tasks.
We follow the WAT18 pre-processing steps\footnote{\url{http://lotus.kuee.kyoto-u.ac.jp/WAT/WAT2018/baseline/dataPreparationJE.html}.} for experiments on En-Ja but use Cabocha\footnote{\url{https://taku910.github.io/cabocha/}.} to tokenize target sentences.
On WMT17, we use \textit{newstest2016} and \textit{newstest2017} as validation and test sets, respectively.

Table~\ref{tab:datasets} lists the final sizes of each data set.

\begin{figure}[t]
	\center
	\includegraphics[width=0.45\columnwidth, trim={0.cm 0cm 0.5cm 3.5cm}, clip]{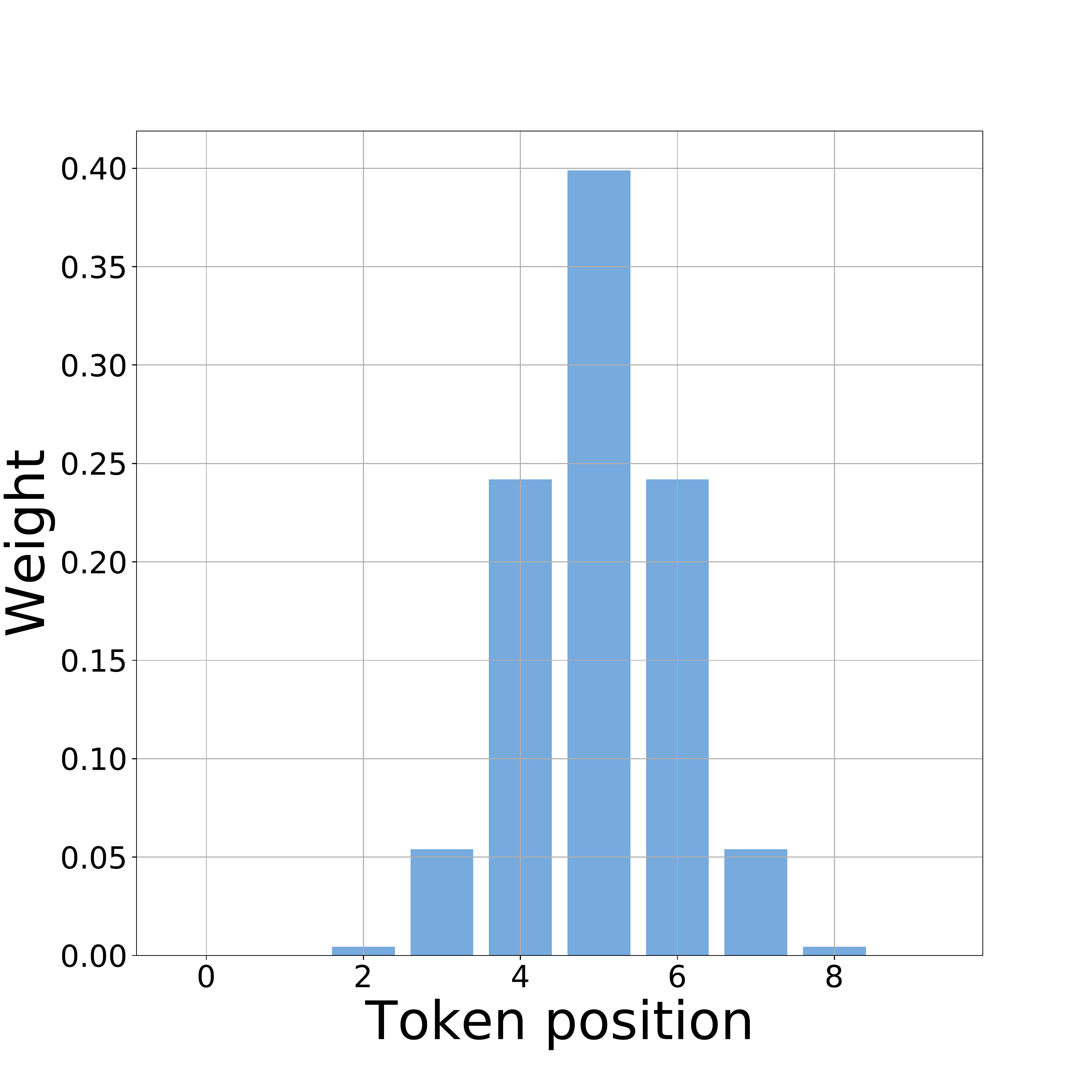}
	\includegraphics[width=0.45\columnwidth, trim={0.cm 0cm 0.5cm 3.5cm}, clip]{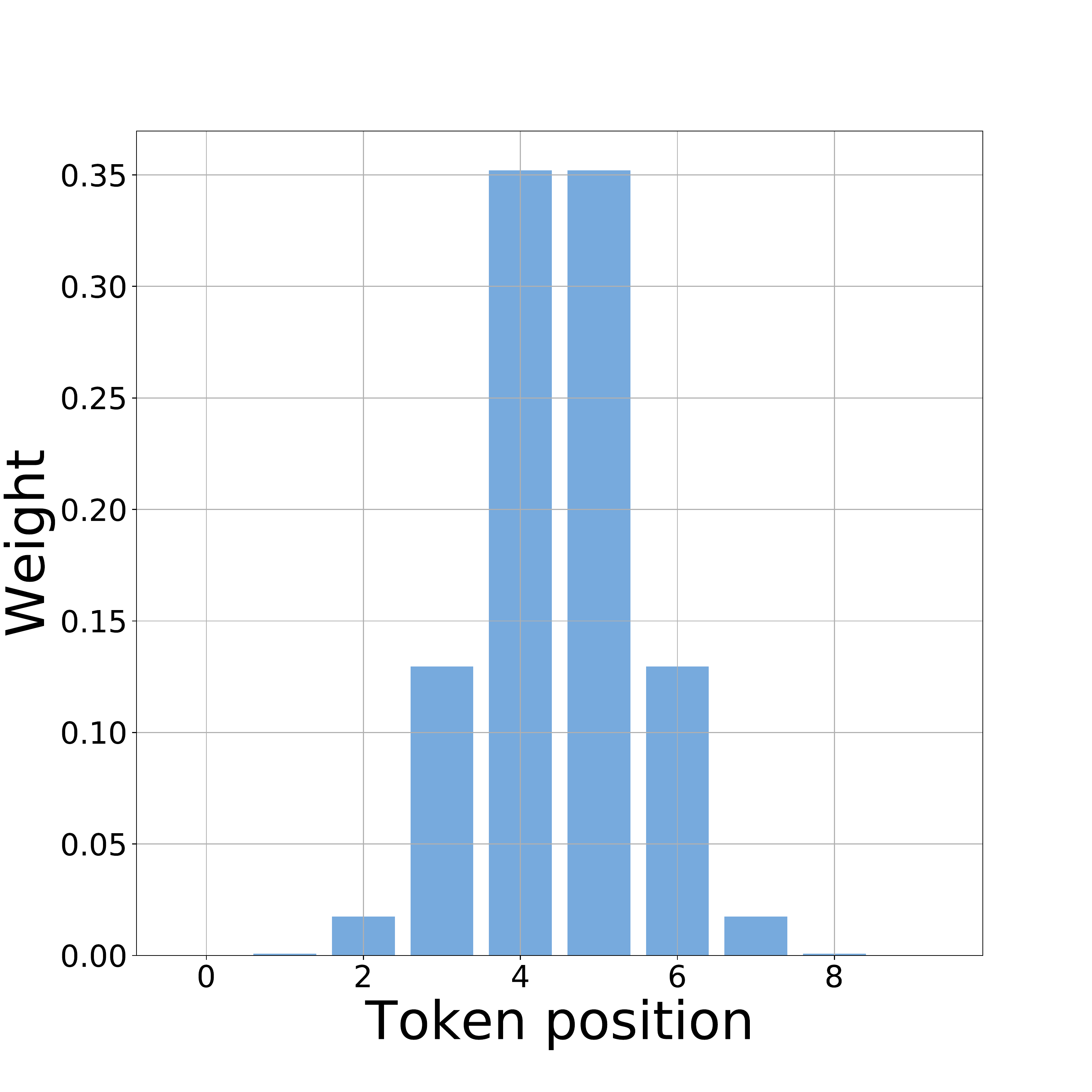}
	\caption{Weights of normal probability density with $\sigma^2=1$ and the means at positions $5$ (left) or $4.5$ (right).}
	\label{fig:normal}
\end{figure}

\begin{table*}[t]
	\centering
	\small
	\setlength\tabcolsep{5pt}
	\begin{tabular}{l|r|r|r|r|r|r}
		\toprule
        \multicolumn{1}{c|}{\textbf{Component}} & \textbf{NC11 En-De} & \textbf{NC11 De-En} & \textbf{WMT18 En-Tr} & \textbf{WMT16 En-De} & \textbf{WMT17 En-De} & \textbf{WAT En-Ja}\\
		\midrule
        Transformer & 22.6 & 23.8 & 12.6 & 29.0 & 31.5 & 42.2 \\
        + data filtering & 22.8 (+0.2) & 24.0 (+0.2) &  & 28.7 (-0.3) &  &  \\
        + \pascal & 23.0 (+0.2) & 24.6 (+0.6) & 13.6 (+1.0) & 29.2 (+0.5) & 31.6 (+0.1) & 43.5 (+1.3) \\
        + parent ignoring & 23.2 (+0.2) &  & 13.7 (+0.1) &  & 32.1 (+0.6) &  \\
		\bottomrule
	\end{tabular}
	\caption{Validation \bleu when incrementally adding each component used by our best-performing models.} \label{tab:ablation}
\end{table*}

\paragraph{Baselines}
We evaluate the impact of syntactic information with the following approaches:
\begin{itemize}[noitemsep,topsep=1pt]
    \item \textbf{Transformer:} We train a base Transformer model as a strong, standard baseline using the hyperparameters in the latest Tensor2Tensor~\cite{tensor2tensor} version ($3$).\\[-12pt]
    \item \textbf{+\sh:} Following~\newcite{sennrich2016linguistic}, we introduce syntactic information in the form of dependency labels in the embedding matrix of the Transformer encoder. More specifically, each token is associated with its dependency label which is first embedded into a vector representation of size $10$ and then used to replace the last $10$ embedding dimensions of the token embedding, ensuring a final size that matches the original one.\\[-12pt]
    \item \textbf{+\multi:} Our implementation of the multi-task approach by~\newcite{currey-heafield-2019-incorporating} where a standard Transformer learns to both parse and translate source sentences. Each source sentence is first duplicated and associated its linearized parse as target sequence.
    To distinguish between the two tasks, a special tag indicating the desired task is prepended and appended to each source sentence.
    Finally, parsing and translation training data is shuffled together.\\[-12pt]
    \item \textbf{+\lisa:} We adapt Linguistically-Informed Self-Attention (LISA;~\citealt{D18-1548}) to NMT. In one attention head $h$, $\mathbf{Q}^h$ and $\mathbf{K}^h$ are computed through a feed-forward layer and the key-query dot product to obtain attention weights is replaced by a bi-affine operator $\mathbf{U}$.
    These attention weights are further supervised to attend to each token's parent by interpreting each row $t$ as the distribution over possible parents for token $t$.
    Here, we extend the authors' approach to BPE by defining the parent of a given token as its first sub-word unit (which represents the root of the parent word).
    The model is trained to maximize the joint probability of translations and parent positions. 
\end{itemize}

\begin{table}[t]
	\centering
	\small
	\begin{tabular}{l|rr}
		\toprule
		\multicolumn{1}{c|}{\textbf{Corpus}} & \textbf{Transformer} & \textbf{+\pascal}\\
		\midrule
        NC11~~~~ En-De & 4,134.1 & 4,188.8 \\
        NC11~~~~ De-En & 4,276.6 & 4,177.4 \\
		WMT18 En-Tr & 3,559.7 & 3,621.1 \\
		\midrule
		WMT16 En-De & 23,186.3 & 23,358.8 \\
		WMT17 En-De & 23,604.1 & 24,083.6 \\
		WAT~~~~~~ En-Ja & 23,005.8 & 23,073.0 \\
		\bottomrule
	\end{tabular}
	\caption{Training times (in seconds) for the Transformer baseline and Transformer+\pascal on each data set. \pascal adds negligible overhead.} \label{tab:times}
\end{table}

\begin{table}[t]
	\centering
	\small
	\setlength\tabcolsep{5pt}
	\begin{tabular}{l|cc|cc}
		\toprule
		\multicolumn{1}{c|}{\textbf{Corpus}} & \textbf{$lr$} & \textbf{$(\beta_1, \beta_2)$} &\textbf{$h_{P}$} &\textbf{$q$}\\
		\midrule
        NC11~~~~ En-De & 0.0007 & (0.9, 0.997) & 2 & 0.4 \\
        NC11~~~~ De-En & 0.0007 & (0.9, 0.997) & 8 & 0.0 \\
		WMT18 En-Tr & 0.0007 & (0.9, 0.980) & 7 & 0.3 \\
		\midrule
		WMT16 En-De & 0.0007 & (0.9, 0.980) & 5 & 0.0 \\
		WMT17 En-De & 0.0007 & (0.9, 0.997) & 7 & 0.3 \\
		WAT~~~~~~ En-Ja & 0.0007 & (0.9, 0.997) & 7 & 0.0 \\
		\bottomrule
	\end{tabular}
	\caption{Hyperparameters for the reported models. $lr$ denotes the maximum learning rate, $(\beta_1, \beta_2)$ are Adam's decay rates, $h_{P}$ is the number of \pascal heads, and $q$ is the parent ignoring probability.} \label{tab:hypers}
\end{table}

\paragraph{Training details}
All experiments are based on the base Transformer architecture and optimized following the learning schedule of~\newcite{vaswani2017attention} with $8,000$ warm-up steps.
Similarly, we use label smoothing $\epsilon_{ls} = 0.1$~\cite{szegedy2016rethinking} during training and employ beam search with a beam size of $4$ and length penalty $\alpha = 0.6$   \cite{wu2016google} at inference time.
We use a batch size of $32K$ tokens and run experiments on a cluster of $4$ machines, each having $4$ Nvidia P100 GPUs.
See Table~\ref{tab:times} for the training times of each experiment.

For each model, we run a small grid search over the hyperparameters and select the ones giving the highest \bleu scores on validation sets (Table~\ref{tab:hypers}).


We use the \textsc{sacreBLEU}~\cite{post-2018-call} tool to compute case-sensitive BLEU scores.\footnote{Signature: BLEU+c.mixed+\#.1+s.exp+tok.13a+v.1.2.12.}
When evaluating En-Ja translations, we follow the procedure employed at WAT by computing BLEU scores after tokenizing target sentences using KyTea.\footnote{\url{http://www.phontron.com/kytea/}.}

Following~\newcite{vaswani2017attention}, we train Transformer-based models for $100K$ steps on large-scale data. 
On small-scale data, we train for $20K$ steps and use a dropout probability $P_{drop} = 0.3$ as they let the Transformer baseline achieve higher performance on this size of data.
For instance, in WMT18 En-Tr, our baseline outperforms the one in~\newcite{currey-heafield-2019-incorporating} by $+3.5$ BLEU.

\section{Analysis}\label{sec:analysis}

\paragraph{Multiplication vs. addition}
In Equation~\eqref{eq:scaling}, we calculated the weighing scores by multiplying the self-attention scores by the distance to the parent token.
Multiplication is, in fact, the standard way to weight values (e.g., the gating mechanism of LSTMs and GRUs). In our case, it introduces sparseness in the attention scores for non-parent tokens.
Moreover, it weights gradients in back-propagation: Let $x$ and $y$ be the attention score and dependency weight, respectively. Consider a loss $l=f(z)$ where $z=xy$ and $dl/dx=df(z)/dz * y$. The attention score receive gradients more on dependent pairs (larger $y$) than non-dependent ones (smaller $y$), which is sound for dependency information. 
In contrast, addition cannot obtain such an effect because it does not affect gradients: $dl/dx=df(z)/dz$ when $z=x+y$.
For completeness, we trained our best NC11 models replacing multiplication by addition. We find that \bleu scores still improve upon the baseline, meaning that our approach is robust, but find them to be slightly lower ($-0.2$) than with multiplication.

\paragraph{Ablation}
We introduced different techniques to improve neural machine translation with syntax information.
Table~\ref{tab:ablation} lists the contribution of each technique, in an incremental fashion, whenever they were used by the models reported in Table~\ref{tab:results}.

While removing sentences whose languages do not match the translation task can lead to better performance (NC11), the precision of the detection tool assumes a major role at large scale.
In WMT16, \texttt{langdetect} removes more than $200K$ sentences and leads to performance losses. It would also drop $19K$ pairs on the clean WAT En-Ja data.

The proposed \pascal mechanism is the component that most improves the performance of the models, achieving up to $+1.0$ and $+1.3$ \bleu on the distant En-Tr and En-Ja pairs, respectively.
With the exception of NC11 En-De, we find parent ignoring useful on the noisier WMT18 En-Tr and WMT17 En-De datasets.
In the former, low-resource case, the benefits of parent ignoring are minimal, but it proves fundamental on the large-scale WMT17 data, where it leads to significant gains when paired with the \pascal mechanism.\footnote{Note that this ablation is obtained by stripping away each component from the best performing models and hence only seeing $+0.1$ for \pascal on WMT17 En-De does not mean that \pascal is not helpful in this task but rather that combining it with parent ignoring gives better performance (our second best model achieved $+0.5$ by using \pascal only).}

Finally, looking at the number of \pascal heads in Table~\ref{tab:hypers}, we notice that most models rely on a large number of syntax-aware heads.
\newcite{raganato-tiedemann-2018-analysis} found that only a few attention heads per layer encoded a significant amount of syntactic dependencies.
Our study shows that the Transformer model can be improved by having more attention heads learn syntactic dependencies.

\end{document}